\documentclass[letterpaper]{article} 
\usepackage[]{aaai25}  
\usepackage{times}  
\usepackage{helvet}  
\usepackage{courier}  
\usepackage[hyphens]{url}  
\usepackage{graphicx} 
\usepackage{natbib}  
\usepackage{caption} 
\usepackage{amsmath}

\usepackage{algorithm}
\usepackage{algorithmic}
\usepackage{amssymb}
\usepackage{array}
\usepackage{multirow}
\usepackage{booktabs}

\usepackage{newfloat}
\usepackage{listings}
\DeclareCaptionStyle{ruled}{labelfont=normalfont,labelsep=colon,strut=off} 
\floatstyle{ruled}
\newfloat{listing}{tb}{lst}{}
\floatname{listing}{Listing}
\title{Beyond Uniform Query Distribution: Key-Driven Grouped Query Attention}
\author {
    Zohaib Khan\equalcontrib,
    Muhammad Khaquan\equalcontrib,
    Omer Tafveez,
    Burhanuddin Samiwala,
    Agha Ali Raza
}
\affiliations {
    Lahore University of Management Sciences\\
    \{24100074, 25100095, 25020254, 25100255, agha.ali.raza\}@lums.edu.pk
}
\nocopyright

\usepackage{bibentry}

\lstdefinestyle{pythonstyle}{
    language=Python,
    basicstyle=\ttfamily\footnotesize, 
    breaklines=true,                   
    numbers=left,                      
    numbersep=5pt,                     
    frame=single,                      
    captionpos=b                       
}

\begin{document}

\maketitle

\begin{abstract}
The Transformer architecture has revolutionized deep learning through its Self-Attention mechanism, which effectively captures contextual information. However, the memory footprint of Self-Attention presents significant challenges for long-sequence tasks. Grouped Query Attention (GQA) addresses this issue by grouping queries and mean-pooling the corresponding key-value heads—reducing the number of overall parameters and memory requirements in a flexible manner without adversely compromising model accuracy. In this work, we introduce enhancements to GQA, focusing on two novel approaches that deviate from the static nature of grouping: \textbf{Key-Distributed GQA (KDGQA)} and \textbf{Dynamic Key-Distributed GQA (DGQA)}, which leverage information from the norms of the key heads to inform query allocation. Specifically, KDGQA looks at the ratios of the norms of the key heads during each forward pass, while DGQA examines the ratios of the norms as they evolve through training. Additionally, we present \textbf{Perturbed GQA (PGQA)} as a case-study, which introduces variability in (static) group formation via subtracting noise from the attention maps. Our experiments with up-trained Vision Transformers, for Image Classification on datasets such as CIFAR-10, CIFAR-100, Food101, and Tiny ImageNet, demonstrate the promise of these variants in improving upon the original GQA through more informed and adaptive grouping mechanisms: specifically ViT-L experiences accuracy gains of up to 8\% when utilizing DGQA in comparison to GQA and other variants. We further analyze the impact of the number of Key-Value Heads on performance, underscoring the importance of utilizing query-key affinities. Code is available on GitHub.\footnote{\url{https://github.com/zohaib-khan5040/key-driven-gqa}}

\end{abstract}

%

\section{Introduction}

Since their inception in 2017, Transformers \citep{vaswani2023attentionneed} have become the state-of-the-art approach in language modeling tasks (\citealp{brown2020languagemodelsfewshotlearners}; \citealp{touvron2023llamaopenefficientfoundation}). More recently, they have been adapted for computer vision tasks such as image classification and segmentation, bypassing the large dependence on convolutions (\citealt{dosovitskiy2021imageworth16x16words}; \citealp{liu2021swintransformerhierarchicalvision}; \citealp{dong2022cswintransformergeneralvision}). The scalability of Transformers has also been empirically established by the scaling laws \citep{hoffmann2022trainingcomputeoptimallargelanguage} which suggest bigger is indeed better in this paradigm. This expansion in model size usually occurs by increasing the number of layers, or the dimensionality of the projections.

\begin{figure}[t]
    \centering
    \includegraphics[width=\columnwidth]{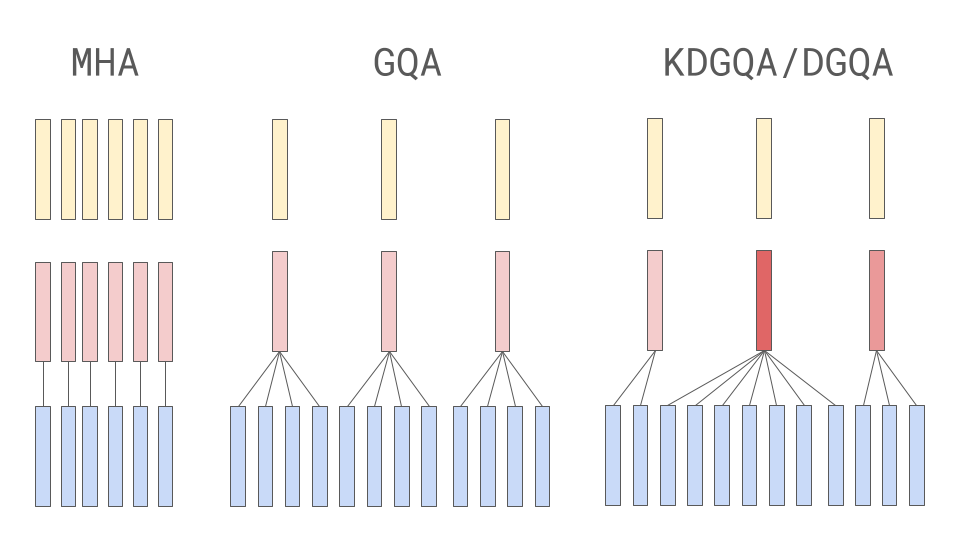}
    \caption{An overview of the relevant mechanisms. MHA associates one query to one key-value head. GQA associates a single key-value head to subgroups of queries such that the group sizes remain constant and the grouping is performed in a static/uniform manner. Our approaches, KDGQA/DGQA, perform the grouping according  to the norms of the key heads in a non-uniform manner - note how the darker keys (indicating higher norms) have more queries assigned to them in comparison to the lighter ones.}
    \label{fig:attention-schemes}
\end{figure}

Increasing model size however comes at the cost of increased memory and computational requirements. Larger models also cause an inference bottleneck since the computational complexity of Self-Attention scales quadratically with number of tokens. These limitations, coupled with the fact that most large transformer based models are at times over-parameterised \citep{hoffmann2022trainingcomputeoptimallargelanguage} has set the stage for various model compression techniques specifically targeting the Self-Attention module.

One of the first popular schemes was Multi-Query Attention \citep{shazeer2019fasttransformerdecodingwritehead} which suggests that in each self-attention block, using only a single key head significantly reduces parameters without causing a sizeable drop in accuracy. Later, Grouped-Query Attention (GQA) was introduced \citep{ainslie2023gqatraininggeneralizedmultiquery} as an interpolation between Multi-Head Attention (MHA) and Multi-Query Attention (MQA). GQA was shown to achieve performance close to MHA, while being nearly as fast as MQA, using only a single key and value head per subgroup of query heads. Since then, there have been multiple approaches that modify the GQA architecture to further mitigate the drop in accuracy (\citealp{javadi2023gqkvaefficientpretrainingtransformers}; \citealp{chinnakonduru2024weightedgroupedqueryattention}; \citealp{chen2024optimisedgroupedqueryattentionmechanism}).

We propose our own extensions onto the GQA architecture which try to soften the accuracy drop by making the grouping mechanism more dynamic and informed in nature, as applied to image classification for Vision Transformers \citep{dosovitskiy2021imageworth16x16words}. Specifically, we explore the possibility and the promise of using a simple heuristic as the $L_2$-norms (as measures of \textit{importance}) of the key heads to guide the group allocation process, during training, in approaches we call \textbf{Key-Distributed GQA (KDGQA)} and \textbf{Dynamic Key-Distributed GQA (DGQA)} - the former limiting its knowledge of the norms to the current forward pass, while the latter examines the ratios as they evolve through training (Figure~\ref{fig:attention-schemes} illustrates the difference in grouping). We also look into \textbf{Perturbed GQA (PGQA)} as an interesting case of introducing Gaussian Noise to the GQA attention maps.

\section{Related Work}

The global nature of Self-Attention, where there are an equal number of queries and keys/values, has made the training and deployment of transformers increasingly computationally expensive. GQA mitigated this to some extent, allowing for forming fixed-size groups of query heads and associating a single key-value head to each group \citep{ainslie2023gqatraininggeneralizedmultiquery}: the paper discusses utilizing existing MHA checkpoints, and converting them to the GQA architecture by mean-pooling the key and value heads associated with each group, respectively. Many works have recently risen that tinker with the connectivity and forward pass dynamics of the Attention mechanism, squeezing out more performance from the model, oftentimes with the same number of parameters.

\textbf{Grouped Query Key Value Attention} \citep{javadi2023gqkvaefficientpretrainingtransformers} emerges as a unifying approach that encapsulates variations like Multi-Key Value Attention (MKVA) and Grouped Key Value Attention (GKVA). While MKVA and GKVA introduce novel strategies by grouping keys and values into distinct groups and sharing queries within these groups, GQKVA partitions queries and key-value pairs, optimizing both the parameter count and computational efficiency. It explores the combinatorial approach of splitting Q matrices into $g$ groups and the K, V matrices into $g_{kv}$ groups, allowing for a flexible number of queries, keys, and values within each partition. This method uses dot-product attention for each unique combination of these groups ($Q_i$, $KV_j$), ensuring distinct outputs across multiple heads. 

\citet{chen2024optimisedgroupedqueryattentionmechanism} propose \textbf{AsymGQA} and investigate the non-uniform allocation of heads by incorporating similar projections into uniformly constructed clusters. For instance, consider a scenario where the set of keys $K$ is partitioned into three groups, each containing two keys. In this approach, the similarity of a randomly selected key is computed with respect to all keys in the other groups. The key is then inserted into the group with the highest similarity. This was also experimented on the activation level, where the output of each layer was used for computing similarity to form clusters of heads, which performed better than Weight-level clustering.

Similarly, \citet{joshi2024qcqaqualitycapacityawaregrouped} propose \textbf{Quality and Capacity-aware GQA (QCQA)}, which uses a minimization loss referred to as the Weight-Sharing Error, which is the distance between the $K$ and $V$ head distributions to minimize the drop in accuracy due to grouping of queries. The approach follows a two-stage search framework, the first forming of the groups of queries $Q$ of each layer independently. This is followed by evaluating groupings of each layer using Weight Sharing Error and KV cache as the fitness functions to drop layers with the largest impact on accuracy. Layers that exhibit significant damage to the accuracy are retained as MHA forms while the others are grouped, finalizing the selection of layers that will preserve these gourpings with non selected layers configured to MHA. The first stage focuses on optimizing groups for each layer individually, while the second stage utilizes the Non-dominated Sorting Genetic Algorithm II to iteratively refine these groupings. 

\textbf{Weighted GQA} \citep{chinnakonduru2024weightedgroupedqueryattention} introduces a learnable scalar or vector parameter to the $K$ and $V$ heads for $(w_{1,k},w_{2,k}..w_{h,k})$ and $(w_{1,v},w_{2,v}..w_{h,v})$. In other words, composite keys and values are formed where each column of $K$ and $V$ is a linear combination of the key and value matrices $K_1, V_1$ through $K_h, V_h$ where the combination is controlled by a specific set of weight vector. The weighted sum of this operation is used to modify $K,V$, which is then used for self-attention.

\section{Method}

This section describes our approaches to diverge from the default uniform nature of the query allocation in Grouped-Query Attention.

\subsection{Key-Distributed GQA (KDGQA)}

This approach was our first attempt at allocating query heads to each group in a more dynamic manner. In KDGQA, rather than uniformly assigning queries to groups, we utilize the magnitudes of the key vectors, measured by their $L_2$-norms as a proxy for their \textit{importance}, to inform the distribution of queries across different groups. We posit that allocating a number of query heads proportional to their \textit{relative importance} (within that layer) is beneficial for model performance.

Given a set of key vectors $\mathbf{K} \in \mathbb{R}^{G \times d_k}$, where $G$ is the number of groups (or key-value heads) and $d_k$ is the dimensionality of the key vectors, the norm of each key vector can be computed as follows:

\begin{equation}
    \mathbf{n}_g = \|\mathbf{K}_g\|_2 = \sqrt{\sum_{i=1}^{d_k} \mathbf{K}_{g,i}^2} \quad \text{for } g = 1, \dots, G
\label{eq:L2norm}
\end{equation}

\noindent Here, $\mathbf{n}_g$ is the norm of the $g$-th key vector $\mathbf{K}_g$.

To ensure that the norms are scaled within a range of $[0, 1]$ for easier interpretation and allocation, we apply min-max scaling to the vector of norms $\mathbf{n} = [n_1, n_2, \dots, n_G]$:

\begin{equation}
    \hat{\mathbf{n}}_g = \frac{\mathbf{n}_g - \min(\mathbf{n})}{\max(\mathbf{n}) - \min(\mathbf{n})}
\label{eq:minmaxscaling}
\end{equation}

Where:
\begin{itemize}
    \item $\hat{\mathbf{n}}_g$ is the scaled norm for the $g$-th group.
    \item $\min(\mathbf{n})$ and $\max(\mathbf{n})$ represent the minimum and maximum values in the vector $\mathbf{n}$.
\end{itemize}

We find that the min-max scaling as described in equation~\ref{eq:minmaxscaling} helps with compressing the norms into a tighter range, amplifying the differences and leading to more non-uniform group sizes, in comparison to directly taking the ratios.

Finally, the number of queries, $Q_g$, allocated to each key head can be determined by normalizing the scaled norms and multiplying by the total number of queries $N_q$:

\begin{equation}
    Q_g = \left\lfloor \hat{\mathbf{n}}_g \cdot \frac{N_q}{\sum_{g=1}^{G} \hat{\mathbf{n}}_g} \right\rfloor
\label{eq:splits_kdgqa}
\end{equation}

Where:
\begin{itemize}
    \item $N_q$ is the total number of queries.
    \item $Q_g$ represents the number of queries allocated to the $g$-th group.
\end{itemize}

KDGQA carries out this procedure for finding the number of query heads to allocate, then forming the groups in a \textit{left-to-right} fashion, during each forward pass, irrespective of training or inference, which causes a small overhead we discuss later (Section~\ref{training_overhead}).

\subsection{Dynamic Key-Distributed GQA (DGQA)}

Dynamic Key-Distributed Grouped Query Attention (DGQA) extends the principles of KDGQA by not just considering the norms of key vectors at one point in time, but by dynamically adjusting these measures of importance during the training process. The core idea of DGQA is to allow the grouping of queries to adapt based on the evolving characteristics of the key vectors as the model learns. This dynamic adaptation aims to more effectively capture and leverage the underlying structure of the data, leading to improved performance in attention mechanisms.

The underlying mechanism relies upon defining a \textit{window size}: a fixed number of iterations at the start of which we update the query allocations and stick with for the duration of the interval. To quantify some measure of how the keys evolve (on the basis of their magnitudes), we propose two approaches:

\begin{enumerate}
    \item \textbf{Difference-based Evolution.} 
    
    We cache the norms of the key heads (as calculated in equation~\ref{eq:L2norm}) at the start of each window/interval $t$, $\mathbf{c}_g^{(t)}$. As training progresses, we quantify the importance of each group by measuring the absolute difference between the cached norm (coming from the previous interval) and the current one:
    \begin{equation}
        \Delta \mathbf{n}_g^{(t)} = \mathbf{n}_g^{(t)} - \mathbf{c}_g^{(t-1)} \quad \text{for} \quad g = 1, \dots, G
        \label{eq:diff_dgqadiff}
    \end{equation}
    where $\Delta \mathbf{n}_g^{(t)}$ represents the change in the norm for the $g$-th key head and the $t$-th interval.

    We then use this difference to form our measure of importance, $\mathbf{d}_g$, for each key head:
    \begin{equation}
        \mathbf{d}_g = \left| \Delta \mathbf{n}_g^{(t)} \right|
        \label{eq:abs_dgqadiff}
    \end{equation}

    The cached values are then updated when the next interval starts:
    \begin{equation}
        \mathbf{c}_g^{(t)} = \mathbf{n}_g^{(t)} \quad \text{for} \quad g = 1, \dots, G
        \label{eq:dgqa_cache}
    \end{equation}

    \item \textbf{Exponential Moving Average-based Evolution.}

    This approach is very similar to the previous one in terms of setting a window-size, using a cache of the norms, and updating the cache. The main difference lies in the measure of importance, $\mathbf{d}_g$: instead of taking an absolute difference of the norms, we take an exponential moving average (EMA) of the two:
    \begin{equation}
        \mathbf{c}_g^{(t)} = \alpha \cdot \mathbf{n}_g^{(t)} + (1 - \alpha) \cdot \mathbf{c}_g^{(t-1)} \quad \text{for} \quad g = 1, \dots, G
    \end{equation}
    where $\alpha$ is a hyperparameter that determines the importance of the current value over the moving average in determining the new average.

    This new cached value would be used for determining the query allocations (note we don't need to take the absolute value here):
    \begin{equation}
        \mathbf{d}_g = \mathbf{c}_g^{(t)} \quad \text{for} \quad g = 1, \dots, G
    \end{equation}
\end{enumerate}

Both approaches simply seek to compute $d_g$ ($g = 1, \dots, G$). Using this, the number of queries allocated in each group, $Q_g$, can be computed in the same fashion as equation~\ref{eq:splits_kdgqa}, taking the place of $\hat{\mathbf{n}}_g$.

\begin{figure}
    \centering
    \includegraphics[width=\columnwidth]{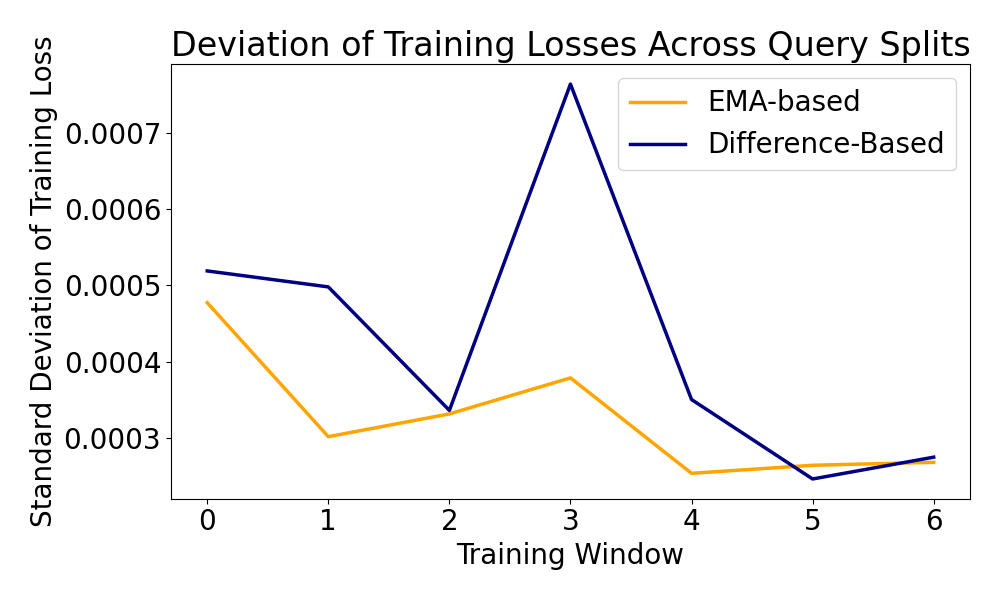}
    \caption{Standard deviation of the EMA-variant incurs fewer spiky, supporting the idea that it is more robust to transient noise.}
    \label{fig:loss-devs}
\end{figure}

Our experiments have established the superiority of the EMA variant over the difference variant: 84\% vs 71\% respectively, for a window size of 300, on CIFAR-100 for 10 epochs, from a converted MHA checkpoint. This can be attributed to the following (Figure \ref{fig:loss-devs}):
\begin{itemize}
    \item \textbf{Smoother Adaptation to Changing Magnitudes}\\ The averaging effect of EMA helps mitigate the influence of outliers or abrupt changes, leading to more robust and reliable head allocations compared to the difference method, especially in scenarios where the norms of keys evolve rapidly.
    \item \textbf{Reduced Sensitivity to Noise}\\ By incorporating past information into the current update, EMA reduces sensitivity to noisy or transient patterns in the data. 
\end{itemize}

For the rest of the paper, note that unless specified otherwise, the DGQA approach refers to the EMA variant. We performed a grid search to find the best performing window size on CIFAR-100, which was found to be 300 - this means that the model's query allocation scheme is updated every 300 training steps, which can be impacted by the size of the dataset. We elected to skip this finer search due to time constraints.

\subsection{Perturbed GQA (PGQA)}

Through experiments, MHA has clear evidence of each head being more similar to itself than the other heads, while GQA fails to maintain this coherency. Oftentimes, the attention output of a head in GQA can be more similar to the other heads in its group, than itself leading to an intra-group similarity bias. Although the scaling laws of large models show little to no performance degradation, the similarity bias contributes to GQA lagging behind MHA in performance. Therefore, we propose Perturbed GQA that removes this similarity bias by introducing a Gaussian Noise matrix in the attention output.

The noise matrix is computed separately for each group $g$ by computing mean $\mu_g$ and standard deviation $\sigma_g$ for each group's attention output $\hat{A}$. A random matrix $\mathbf{R}_g$ is generated of the same shape as $\hat{A}_g$ and is normalized to match the statistics of $\hat{A}_g$ as follows:

\begin{equation}
    \mathbf{Gaussian}_g = \sigma_g * \frac{(\mathbf{R}_g - \mu(\mathbf{R}_g))}{\sigma(\mathbf{R}_g)} + \mu_g \quad \text{for} \quad g=1, \dots, G
\end{equation}
where $\mathbf{Gaussian}_g$ is the Gaussian Noise matrix computed for each group. To maintain sparsity and avoid disrupting the attention patterns, we set the diagonal elements of the matrix to zero. 

The obtained Gaussian Noise Matrix is subtracted from the attention output of each group. Meanwhile, the group allocation is carried with the same fashion from \textit{left-right} as GQA.

\begin{figure}
    \centering
    \includegraphics[width=3.5in, height=2in]{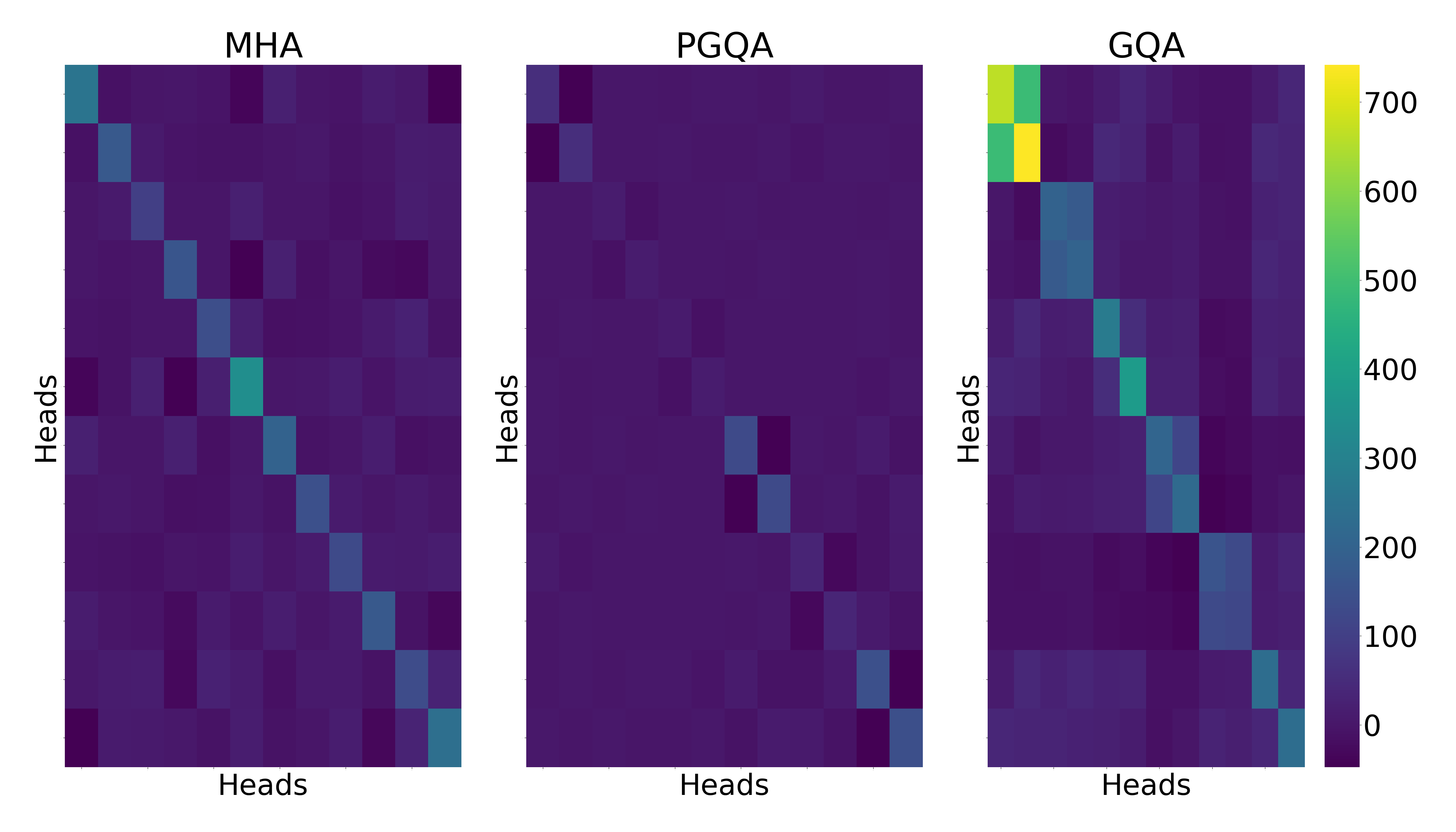}
    \caption{PGQA mitigates similarity bias from GQA, however it wipes out self-similarity patterns seen in attention map of MHA.}
    \label{fig:heatmap}
\end{figure}

Figure \ref{fig:heatmap} shows that PGQA helps us mitigate the intra-group similarity bias. However, this bias mitigation also causes the similarity of each head with itself to diminish. In Appendix~\ref{appendix_head_sim}, we also discuss the head similarity patterns observed in DGQA which approximates a weighted average of the heat maps from MHA and GQA.

\section{Experiments}

\subsection{Models and Datasets}

We elected to use Vision Transformers \citep{dosovitskiy2021imageworth16x16words} in our experiments, following an experimentation scheme similar to \citet{javadi2023gqkvaefficientpretrainingtransformers}. This choice is favorable since pre-trained Vision Transformers come in three model sizes, which can easily fit in consumer-level hardware. Approximate values for the model sizes, and the number of heads they were configured to train with, are shown in Table~\ref{tab:models}.

\begin{table*}[h]
\centering
\begin{tabular}{lccc}
\textbf{Model} & \textbf{Number of Parameters (M)} & \textbf{Size (MB)} & \textbf{Number of Heads} \\ \hline
ViT-S & 30 & 77 & 6 \\ 
ViT-B & 75 & 300 &  12 \\ 
ViT-L & 300 & 1060 & 16 \\ 
\end{tabular}
\caption{Model parameters, sizes, and number of heads. Note that (1) the number of heads come from pretrained model's configuration, as it is a hyperparameter that is set before pretraining, and (2) the Size in MB comes from our uptraining experiments on ImageNet-1k. }
\label{tab:models}
\end{table*}

This flexibility allows for a more robust evaluation when paired with our choice of datasets: CIFAR-10, CIFAR-100, Food101, and Tiny ImageNet (\citet{Krizhevsky09learningmultiple}; \citet{bossard14}; \citet{tiny-imagenet}).

\subsection{Checkpoint Conversion and Uptraining}

The "uptraining" scheme proposed in the GQA paper by \citet{ainslie2023gqatraininggeneralizedmultiquery} describes adapting pre-trained MHA checkpoints to the GQA architecture: this involves the construction of each group's key and value heads by mean-pooling all the original heads within that group, followed by continued pre-training to close the gap between GQA and MHA. We perform the checkpoint conversion using MHA checkpoints of ViT from the \texttt{timm} library \citep{rw2019timm}. 

With the converted model, we perform up-training in a fashion similar to what was described in the original ViT paper. We use the AdamW optimizer with a learning rate of $1 \times 10^{-4}$, with $\beta_1=0.9$, $\beta_2=0.999$, $\epsilon=1\times 10^{-8}$ and a weight decay of $0.01$. We train for $1$ epoch on the ImageNet-1k dataset \citep{imagenet15russakovsky}, with over 1.3M training images. This is carried out with a batch size of $32$ on a single NVIDIA GeForce RTX 4090 24GB GPU. Note that since all models have the same number of parameters, keeping the size (small, base, large) and dataset fixed, the only difference in uptraining FLOPs stem from the additional computations in finding the number of queries to allocate to each group.

For up-training, we apply a random augmentation to the images before they are resized to an image size of $(224,224,3)$. This is to mitigate overfitting which is a big challenge when training ViTs - augmentations are preferable in comparison to regularization strategies as discussed by \citet{steiner2022trainvitdataaugmentation}.

\begin{figure}
    \centering
    \includegraphics[width=\columnwidth]{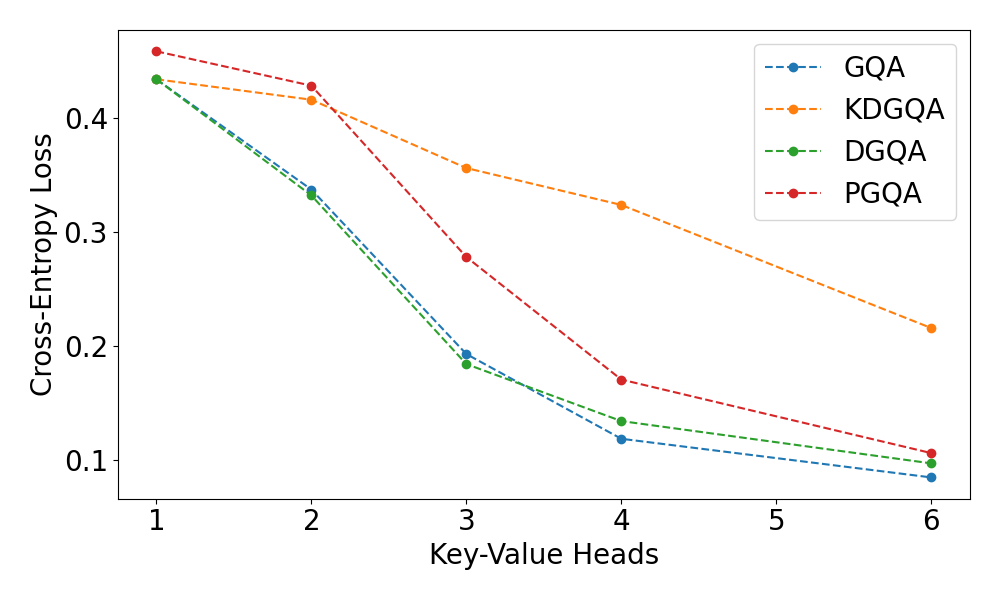}
    \caption{Scaling the number of Key-Value heads, and how it impacts the loss. Each data point represents a variant on ViT-B, with 12 heads, that was finetuned on CIFAR-100 for 5 epochs, without any uptraining for the sake of time. It is evident that the models perform better when they have the maximal number of Key-Value heads.}
    \label{fig:num_kv_scaling}
\end{figure}

Note that for uptraining and for fine-tuning, we elected to pit the best versions of our variants against the best version of GQA, i.e. with the maximum number of key-value heads. This allows for maximally narrowing the gap between GQA (plus its variants) and MHA: with $H$ heads for a given model, we go with $\frac{H}{2}$ key-value heads for fairness, since that is the ceiling of what is allowed with GQA. We additionally examine the impact on varying the number of key-value heads on performance in Figure~\ref{fig:num_kv_scaling}.

\subsection{Fine-tuning}

To evaluate our models, we load in the model checkpoints from the uptraining phase. It is important to note that \textit{all models undergo the same number of training steps}, so there is no advantage given to our variants despite the modifications made. \textit{No extra opportunity is given to our variants to recover model performance.}

Each model is fine-tuned for $5$ epochs. The optimizer, hardware, augmentation strategy, and batch size is the same as the uptraining runs, with the exception of the learning rate which is set to $1 \times 10^{-5}$ for fine-tuning.

\section{Results and Discussion}

\subsection{Evaluation}

The results from the uptraining runs are shown in Table~\ref{tab:results_uptraining}. Our core results are shown in Table~\ref{tab:results_1e-5_main}, where we uptrain each model and variant, and Table~\ref{tab:results_1e-5_gqackpt}, where we use the uptrained GQA checkpoint and only introduce our modifications during fine-tuning. For these tables, GQA and its variants have the following configuration: ViT-S uses 3 key-value heads, ViT-B uses 6 key-value heads, and ViT-L uses 8 key-value heads. Our benchmark is the GQA variant, and the MHA variants' performance have been provided for reference.

\begin{table}[h]
\centering
\begin{tabular}{ccc}
    \toprule
    \textbf{Model} & \textbf{Variant} & \textbf{Accuracy} \\
     
     \midrule

    \multirow{4}{*}{ViT-S} 
     & GQA & 65.95 \\ 
     & KDGQA & 51.85 \\ 
     & DGQA & 65.66 \\ 
     & PGQA & 59.55 \\

     \midrule

    \multirow{4}{*}{ViT-B} 
     & GQA & 48.43  \\ 
     & KDGQA & 43.69 \\ 
     & DGQA & 45.53  \\ 
     & PGQA & 44.59 \\

     \midrule
     
    \multirow{4}{*}{ViT-L}
     & GQA & 52.49  \\ 
     & KDGQA & 37.21 \\ 
     & DGQA & \textbf{59.13} \\ 
     & PGQA & 49.94 \\
    
     \bottomrule
\end{tabular}
\caption{Validation accuracies after 1 epoch of uptraining on ImageNet-1k. Models outperforming GQA in its category are highlighted in \textbf{bold}.}
\label{tab:results_uptraining}
\end{table}

\begin{table*}[h!]
\centering
\begin{tabular}{cccccc}
    \toprule
    
    \multirow{3}{*}{\textbf{Model}} & \multirow{3}{*}{\textbf{Variant}} & \multicolumn{4}{c}{\textbf{Accuracy}} \\
     &  & \textbf{CIFAR-10} & \textbf{CIFAR-100} & \textbf{Food101} & \textbf{Tiny ImageNet} \\
     
     \midrule

    \multirow{5}{*}{ViT-S} 
     & MHA & 98.42 & 90.94 & 87.29 & 86.51 \\ 
     & GQA & 97.27 & 84.70 & 80.95 & 80.23 \\ 
     & KDGQA & 94.05 & 71.32 & 72.04 & 63.66 \\ 
     & DGQA & 97.27 & 84.55 & 80.77 & 80.16 \\ 
     & PGQA & 95.81 & 78.42 & 76.55 & 73.81 \\

     \midrule

    \multirow{5}{*}{ViT-B} 
     & MHA & 98.67 & 91.91 & 90.17 & 91.04 \\ 
     & GQA & 93.32 & 71.78 & 72.42 & 62.87 \\ 
     & KDGQA & 91.33 & 67.84 & 68.01 & 57.90 \\ 
     & DGQA & 91.95 & 69.63 & 70.59 & 59.55 \\ 
     & PGQA & 91.84 & 67.86 & 68.31 & 59.27 \\

     \midrule
     
    \multirow{5}{*}{ViT-L} 
     & MHA & 99.10 & 94.05 & 91.94 & 92.60 \\ 
     & GQA & 94.81 & 76.41 & 74.57 & 67.73 \\ 
     & KDGQA & 88.87 & 64.50 & 62.90 & 52.49 \\ 
     & DGQA & \textbf{96.36} & \textbf{81.67} & \textbf{79.93} & \textbf{75.83} \\ 
     & PGQA & 93.21 & 75.03 & 73.33 & 66.78 \\
    
     \bottomrule
\end{tabular}
\caption{Test accuracies of the finetuning runs, \underline{post-uptraining for each variant}. The models outperforming GQA in its category are highlighted in \textbf{bold}.}
\label{tab:results_1e-5_main}
\end{table*}

\begin{table*}[h!]
\centering
\begin{tabular}{cccccc}
    \toprule
    
    \multirow{3}{*}{\textbf{Model}} & \multirow{3}{*}{\textbf{Variant}} & \multicolumn{4}{c}{\textbf{Accuracy}} \\
     &  & \textbf{CIFAR-10} & \textbf{CIFAR-100} & \textbf{Food101} & \textbf{Tiny ImageNet} \\
     
     \midrule

    \multirow{5}{*}{ViT-S} 
     & MHA & 98.42 & 90.94 & 87.29 & 86.51 \\ 
     & GQA & 97.27 & 84.70 & 80.95 & 80.23 \\ 
     & KDGQA & 93.57 & 64.34 & 71.02 & 65.83 \\ 
     & DGQA & 97.09 & 84.54 & 80.81 & 80.03 \\ 
     & PGQA & 95.81 & 78.68 & 75.44 & 75.79 \\

     \midrule

    \multirow{5}{*}{ViT-B} 
     & MHA & 98.67 & 91.91 & 90.17 & 91.04 \\ 
     & GQA & 93.32 & 71.78 & 72.42 & 62.87 \\ 
     & KDGQA & 90.36 & 65.64 & 65.99 & 56.42 \\ 
     & DGQA & 92.80 & 70.17 & 71.70 & 61.42 \\ 
     & PGQA & 91.85 & 68.19 & 68.73 & 59.75 \\

     \midrule
     
    \multirow{5}{*}{ViT-L} 
     & MHA & 99.10 & 94.05 & 91.94 & 92.60 \\ 
     & GQA & 94.81 & 76.41 & 74.57 & 67.73 \\ 
     & KDGQA & 86.69 & 63.34 & 67.18 & 61.67 \\ 
     & DGQA & 93.77 & 75.83 & 73.79 & 66.04 \\ 
     & PGQA & 94.19 & 73.66 & 71.91 & 66.84 \\
    
     \bottomrule
\end{tabular}
\caption{Test accuracies of the finetuning runs, \underline{post-uptraining using only the GQA checkpoint}.}
\label{tab:results_1e-5_gqackpt}
\end{table*}

From Table~\ref{tab:results_1e-5_main}, the most important note is that DGQA consistently outperforms the GQA benchmark when using an uptrained ViT-L architecture, and by rather \textit{large} margins, by up to $8\%$ as is the case for Tiny ImageNet. 

It is very interesting to note that for ViT-S and ViT-B, DGQA always underperforms in comparison to GQA - we wished to include these failed experiments in our evaluation as to highlight how much of an impact the number of key-value heads and size of the model impact the change in performance. Since DGQA was designed as an improvement over KDGQA, it is no surprise that the latter performs poorly in our final evaluation, as was the same pattern in preliminary experiments.

Another interesting pattern is in the complexity of the classification task: if we hone in on ViT-L, we find (approximately) that there is a $1.5\%$ gain in accuracy for CIFAR-10, $5\%$ gains for CIFAR-100 and Food101 (note they have 100 and 101 classes respectively), and $8\%$ for Tiny ImageNet (which has 200 classes). This, in tandem with the note from the previous paragraph, lead us to postulate that the true benefits of new approaches, derived from branching off a pretrained base, only come forward on larger scales, i.e. with larger models and bigger datasets.

While PGQA may underperform the benchmark in our evaluations, it follows the same pattern above (narrower margin when scaling up), and presents an intriguing case study. Despite its subpar performance, PGQA provides valuable insights into the (lack of) resilience of attention mechanisms. It suggests that even small perturbations can impact attention distributions, especially when the model is already tuned to operate in a specific manner. This finding underlines the importance of extensive uptraining when making such modifications to pretrained models, as the forward pass dynamics are significantly altered, necessitating additional adjustments to recover model performance.

From Table~\ref{tab:results_1e-5_gqackpt}, it can be seen that our variants perform worse on an absolute scale, irrespective of comparisons to GQA. This highlights that special uptraining must be conducted for variants that are modifications to GQA, coming from conversions to MHA checkpoints.

\subsection{Inference Time}
\label{training_overhead}

Table~\ref{tab:inference_time} presents a comparison of inference time for base models with GQA serving as the baseline. PGQA exhibits the highest increase in inference time by $3.31\%$, while DGQA and KDGQA show minimal increases of $0.45\%$ and $0.21\%$ respectively.

\begin{table}[h]
    \centering
    \begin{tabular}{ccc}
    \toprule
    \textbf{Model} & \textbf{Inference Time in Seconds} & \textbf{\% Difference} \\
    \midrule
    
    GQA & 0.192415  & - \\ 
    DGQA & 0.193276 & 0.4477\%\\ 
    KDGQA & 0.192812 & 0.2064\% \\
    PGQA & 0.198783 & 3.3098\% \\ 
    
    \end{tabular}
    \caption{Inference time comparison between ViT-B variants averaged across batch size of 288, single NVIDIA GeForce RTX 4090 24GB GPU.}
    \label{tab:inference_time}
\end{table}

\subsection{Sensitivity to Training}

Our results clearly show that \textit{asymmetric groupings} and \textit{activation informed} allocations can help improve performance. However, there are many considerations when aiming to improve mechanisms for existing models.

From the previous subsection, we again point out that the scale/size of the dataset and model play a significant role in determining whether the variants will have the capacity to outperform the benchmark. With larger models, there are a larger number of heads, which means there is a bigger ceiling on the number of key-value heads we can tinker with. A larger number of key-value heads allows informed-allocation mechanisms to really shine, since having too few key-value heads implies there is a much narrower margin for error in forming groups. This would likely be less of an issue when applying to Transformers for natural-language tasks since many popular benchmarks and models are generally very large in scale \citep{hoffmann2022trainingcomputeoptimallargelanguage}.

This is not an isolated occurence of the sensitivity of a new approach to the scale of the model and dataset. \citet{dosovitskiy2021imageworth16x16words} make similar notes in their paper on the Vision Transformer when discussing the pretraining data requirements: ViT-L underperforms in comparison to ViT-B on ImageNet-1k, performs similarly on ImageNet-21k, and surpasses on JFT-300M. We hypothesize that this pattern could be extrapolated to our study in that much more calibration is required for activation-informed grouping approaches to surpass the static grouping approach.

It is also worth pointing out again that KDGQA and DGQA have the same fundamental idea, but the latter takes a slightly different perspective in one step of the computation. Despite this small nuance, it is very clear that DGQA is significantly better than KDGQA, and that too only the EMA variant.

Another challenge we experienced was that of the learning rates. We found that a learning rate of $1 \times 10^{-3}$ was too high for our variants specifically, breaking their performance on all datasets. A lower, and more popular, learning rate of $3 \times 10^{-4}$ performed better for PGQA, but still broke DGQA. We finally started seeing reasonable results with $1 \times 10^{-4}$, and settled on using $1 \times 10^{-5}$ (see Appendix~\ref{appendix_learningrate}).

The final ingredient in the recipe for training a variant of a (converted) pretrained model was the \textbf{uptraining dataset}. For preliminary experiments, we tinkered with training GQA and its variants 
\begin{enumerate}
    \item From scratch,
    \item Converting from a MHA checkpoint but not uptraining,
    \item Uptraining on CINIC-10 for 10 epochs,
    \item Uptraining on ImageNet-1k for 1 epoch.
\end{enumerate}

We found that the results using our variants were rather poor when the models were not uptrained properly: specifically when we used a modified version of CINIC-10 \citep{darlow2018cinic10imagenetcifar10}, which had 270000 training images. The lack of variety and the raw low-resolution made this a poor choice, which became evident when we uptrained on ImageNet-1k for just one epoch. Though the number of training steps in the fourth setup was less than a half that of the third, the impact of the quality of uptraining was significant since that is when we began to see our variants outperform the GQA benchmark. Due to time and resource constraints, we were unable to experiment with uptraining on ImageNet-21k or JFT-300M, or longer calibration on ImageNet-1k.

We explore this idea in more depth in Appendix~\ref{appendix_uptraining}.

\section{Conclusion}

We introduce KDGQA, DGQA, and PGQA: strategies for moving past the static grouping strategy as was found in GQA. We find that DGQA, being a refined version of KDGQA, significantly outperforms uptrained GQA on ViT-L, and shows larger gains in performance for more sophisticated datasets. This shows the promise of calibrating for more informed and adaptive grouping mechanisms. 

\section{Future Works}

Since our experiments demonstrated the viability of improving the performance of Transformers using GQA through a more informed grouping mechanism and diverging from uniform allocations, the first step would be to extend these to Decoders. With the popularity of Natural Language tasks nowadays, it would be an interesting series of experiments to tinker with LLMs that have billions of parameters and deal with tasks that have a much higher intrinsic dimensionality, such as language modeling. Based on our experiences, we believe that this combination could lead to even larger performance gains with the usage of informed grouping mechanisms.

Even within the model-agnostic framework, there are interesting directions we plan to explore. Firstly, we note that using the norms of the Keys, or manipulating the attention maps, still does not actually allow the activations to \textit{communicate} with each other. Specifically, we plan to explore whether it is possible to leverage the affinities between the queries and keys, and the keys and values, to improve upon the grouping mechanism even more. We could also divert from just activation-informed approaches, and look at weight-informed approaches, to dictate the allocations independently of the input. These are directions in the same neighborhood as \citet{chen2024optimisedgroupedqueryattentionmechanism}, but we wish to take less of a stochastic approach to finding optimal groupings and give both the variants and the benchmark the same number of training steps for a fair evaluation.
\bibliography{aaai25}

\begin{thebibliography}{20}
\providecommand{\natexlab}[1]{#1}

\bibitem[{Ainslie et~al.(2023)Ainslie, Lee-Thorp, de~Jong, Zemlyanskiy, Lebrón, and Sanghai}]{ainslie2023gqatraininggeneralizedmultiquery}
Ainslie, J.; Lee-Thorp, J.; de~Jong, M.; Zemlyanskiy, Y.; Lebrón, F.; and Sanghai, S. 2023.
\newblock GQA: Training Generalized Multi-Query Transformer Models from Multi-Head Checkpoints.
\newblock arXiv:2305.13245.

\bibitem[{Bossard, Guillaumin, and Van~Gool(2014)}]{bossard14}
Bossard, L.; Guillaumin, M.; and Van~Gool, L. 2014.
\newblock Food-101 -- Mining Discriminative Components with Random Forests.
\newblock In \emph{European Conference on Computer Vision}.

\bibitem[{Brown et~al.(2020)Brown, Mann, Ryder, Subbiah, Kaplan, Dhariwal, Neelakantan, Shyam, Sastry, Askell, Agarwal, Herbert-Voss, Krueger, Henighan, Child, Ramesh, Ziegler, Wu, Winter, Hesse, Chen, Sigler, Litwin, Gray, Chess, Clark, Berner, McCandlish, Radford, Sutskever, and Amodei}]{brown2020languagemodelsfewshotlearners}
Brown, T.~B.; Mann, B.; Ryder, N.; Subbiah, M.; Kaplan, J.; Dhariwal, P.; Neelakantan, A.; Shyam, P.; Sastry, G.; Askell, A.; Agarwal, S.; Herbert-Voss, A.; Krueger, G.; Henighan, T.; Child, R.; Ramesh, A.; Ziegler, D.~M.; Wu, J.; Winter, C.; Hesse, C.; Chen, M.; Sigler, E.; Litwin, M.; Gray, S.; Chess, B.; Clark, J.; Berner, C.; McCandlish, S.; Radford, A.; Sutskever, I.; and Amodei, D. 2020.
\newblock Language Models are Few-Shot Learners.
\newblock arXiv:2005.14165.

\bibitem[{Chen et~al.(2024)Chen, Zhang, Gao, Mullins, Constantinides, and Zhao}]{chen2024optimisedgroupedqueryattentionmechanism}
Chen, Y.; Zhang, C.; Gao, X.; Mullins, R.~D.; Constantinides, G.~A.; and Zhao, Y. 2024.
\newblock Optimised Grouped-Query Attention Mechanism for Transformers.
\newblock arXiv:2406.14963.

\bibitem[{Chinnakonduru and Mohapatra(2024)}]{chinnakonduru2024weightedgroupedqueryattention}
Chinnakonduru, S.~S.; and Mohapatra, A. 2024.
\newblock Weighted Grouped Query Attention in Transformers.
\newblock arXiv:2407.10855.

\bibitem[{Darlow et~al.(2018)Darlow, Crowley, Antoniou, and Storkey}]{darlow2018cinic10imagenetcifar10}
Darlow, L.~N.; Crowley, E.~J.; Antoniou, A.; and Storkey, A.~J. 2018.
\newblock CINIC-10 is not ImageNet or CIFAR-10.
\newblock arXiv:1810.03505.

\bibitem[{Dong et~al.(2022)Dong, Bao, Chen, Zhang, Yu, Yuan, Chen, and Guo}]{dong2022cswintransformergeneralvision}
Dong, X.; Bao, J.; Chen, D.; Zhang, W.; Yu, N.; Yuan, L.; Chen, D.; and Guo, B. 2022.
\newblock CSWin Transformer: A General Vision Transformer Backbone with Cross-Shaped Windows.
\newblock arXiv:2107.00652.

\bibitem[{Dosovitskiy et~al.(2021)Dosovitskiy, Beyer, Kolesnikov, Weissenborn, Zhai, Unterthiner, Dehghani, Minderer, Heigold, Gelly, Uszkoreit, and Houlsby}]{dosovitskiy2021imageworth16x16words}
Dosovitskiy, A.; Beyer, L.; Kolesnikov, A.; Weissenborn, D.; Zhai, X.; Unterthiner, T.; Dehghani, M.; Minderer, M.; Heigold, G.; Gelly, S.; Uszkoreit, J.; and Houlsby, N. 2021.
\newblock An Image is Worth 16x16 Words: Transformers for Image Recognition at Scale.
\newblock arXiv:2010.11929.

\bibitem[{Hoffmann et~al.(2022)Hoffmann, Borgeaud, Mensch, Buchatskaya, Cai, Rutherford, de~Las~Casas, Hendricks, Welbl, Clark, Hennigan, Noland, Millican, van~den Driessche, Damoc, Guy, Osindero, Simonyan, Elsen, Rae, Vinyals, and Sifre}]{hoffmann2022trainingcomputeoptimallargelanguage}
Hoffmann, J.; Borgeaud, S.; Mensch, A.; Buchatskaya, E.; Cai, T.; Rutherford, E.; de~Las~Casas, D.; Hendricks, L.~A.; Welbl, J.; Clark, A.; Hennigan, T.; Noland, E.; Millican, K.; van~den Driessche, G.; Damoc, B.; Guy, A.; Osindero, S.; Simonyan, K.; Elsen, E.; Rae, J.~W.; Vinyals, O.; and Sifre, L. 2022.
\newblock Training Compute-Optimal Large Language Models.
\newblock arXiv:2203.15556.

\bibitem[{Javadi et~al.(2023)Javadi, Ahmed, Hajimolahoseini, Ataiefard, Hassanpour, Asani, Wen, Awad, Liu, and Liu}]{javadi2023gqkvaefficientpretrainingtransformers}
Javadi, F.; Ahmed, W.; Hajimolahoseini, H.; Ataiefard, F.; Hassanpour, M.; Asani, S.; Wen, A.; Awad, O.~M.; Liu, K.; and Liu, Y. 2023.
\newblock GQKVA: Efficient Pre-training of Transformers by Grouping Queries, Keys, and Values.
\newblock arXiv:2311.03426.

\bibitem[{Joshi et~al.(2024)Joshi, Laddha, Sinha, Omer, and Subramoney}]{joshi2024qcqaqualitycapacityawaregrouped}
Joshi, V.; Laddha, P.; Sinha, S.; Omer, O.~J.; and Subramoney, S. 2024.
\newblock QCQA: Quality and Capacity-aware grouped Query Attention.
\newblock arXiv:2406.10247.

\bibitem[{Krizhevsky(2009)}]{Krizhevsky09learningmultiple}
Krizhevsky, A. 2009.
\newblock Learning multiple layers of features from tiny images.
\newblock Technical report.

\bibitem[{Liu et~al.(2021)Liu, Lin, Cao, Hu, Wei, Zhang, Lin, and Guo}]{liu2021swintransformerhierarchicalvision}
Liu, Z.; Lin, Y.; Cao, Y.; Hu, H.; Wei, Y.; Zhang, Z.; Lin, S.; and Guo, B. 2021.
\newblock Swin Transformer: Hierarchical Vision Transformer using Shifted Windows.
\newblock arXiv:2103.14030.

\bibitem[{mnmoustafa(2017)}]{tiny-imagenet}
mnmoustafa, M.~A. 2017.
\newblock Tiny ImageNet.

\bibitem[{Russakovsky et~al.(2015)Russakovsky, Deng, Su, Krause, Satheesh, Ma, Huang, Karpathy, Khosla, Bernstein, Berg, and Fei-Fei}]{imagenet15russakovsky}
Russakovsky, O.; Deng, J.; Su, H.; Krause, J.; Satheesh, S.; Ma, S.; Huang, Z.; Karpathy, A.; Khosla, A.; Bernstein, M.; Berg, A.~C.; and Fei-Fei, L. 2015.
\newblock {ImageNet Large Scale Visual Recognition Challenge}.
\newblock \emph{International Journal of Computer Vision (IJCV)}, 115(3): 211--252.

\bibitem[{Shazeer(2019)}]{shazeer2019fasttransformerdecodingwritehead}
Shazeer, N. 2019.
\newblock Fast Transformer Decoding: One Write-Head is All You Need.
\newblock arXiv:1911.02150.

\bibitem[{Steiner et~al.(2022)Steiner, Kolesnikov, Zhai, Wightman, Uszkoreit, and Beyer}]{steiner2022trainvitdataaugmentation}
Steiner, A.; Kolesnikov, A.; Zhai, X.; Wightman, R.; Uszkoreit, J.; and Beyer, L. 2022.
\newblock How to train your ViT? Data, Augmentation, and Regularization in Vision Transformers.
\newblock arXiv:2106.10270.

\bibitem[{Touvron et~al.(2023)Touvron, Lavril, Izacard, Martinet, Lachaux, Lacroix, Rozière, Goyal, Hambro, Azhar, Rodriguez, Joulin, Grave, and Lample}]{touvron2023llamaopenefficientfoundation}
Touvron, H.; Lavril, T.; Izacard, G.; Martinet, X.; Lachaux, M.-A.; Lacroix, T.; Rozière, B.; Goyal, N.; Hambro, E.; Azhar, F.; Rodriguez, A.; Joulin, A.; Grave, E.; and Lample, G. 2023.
\newblock LLaMA: Open and Efficient Foundation Language Models.
\newblock arXiv:2302.13971.

\bibitem[{Vaswani et~al.(2023)Vaswani, Shazeer, Parmar, Uszkoreit, Jones, Gomez, Kaiser, and Polosukhin}]{vaswani2023attentionneed}
Vaswani, A.; Shazeer, N.; Parmar, N.; Uszkoreit, J.; Jones, L.; Gomez, A.~N.; Kaiser, L.; and Polosukhin, I. 2023.
\newblock Attention Is All You Need.
\newblock arXiv:1706.03762.

\bibitem[{Wightman(2019)}]{rw2019timm}
Wightman, R. 2019.
\newblock PyTorch Image Models.
\newblock \url{https://github.com/rwightman/pytorch-image-models}.

\end{thebibliography}
\appendix
\newpage

\section{Learning Rates}
\label{appendix_learningrate}

Following the notes from our primary evaluation, we wanted to show the impact of learning rates on the models and the variants. A learning rate of $1 \times 10^{-4}$ was where training began to stabilize across all models and variants - we found that lower learning rates (up to $1 \times 10^{-5}$) were more beneficial for fine-tuning.

Table~\ref{tab:results_1e-4_var_specific} shows the results from the same setup as Table~\ref{tab:results_1e-5_main}, but with a higher learning rate. We find that some of the ViT-S variants only slightly outperform the GQA benchmark, though not by significant margins, and that DGQA outperforms GQA consistently when using ViT-L.

If numbers are compared across these two tables, it can be found that the models perform better with lower learning rates - the gains in performance by DGQA are also more significant, indicating that these models are much more sensitive to the training setup.

\begin{table*}[h!]
\centering
\begin{tabular}{cccccc}
    \toprule
    
    \multirow{3}{*}{\textbf{Model}} & \multirow{3}{*}{\textbf{Variant}} & \multicolumn{4}{c}{\textbf{Accuracy}} \\
     &  & \textbf{CIFAR-10} & \textbf{CIFAR-100} & \textbf{Food101} & \textbf{Tiny ImageNet} \\
     
     \midrule

    \multirow{5}{*}{ViT-S} 
     & MHA & 96.29 & 85.87 & 84.93 & 78.61 \\ 
     & GQA & 95.73 & 83.77 & 82.22 & 76.38 \\ 
     & KDGQA & 93.14 & 76.53 & 77.78 & 66.64 \\ 
     & DGQA & \textbf{96.10} & 83.57 & \textbf{82.30} & 76.16 \\ 
     & PGQA & 95.19 & 81.03 & 80.03 & 73.02 \\

     \midrule

    \multirow{5}{*}{ViT-B} 
     & MHA & 95.08 & 81.86 & 81.13 & 73.04 \\ 
     & GQA & 92.80 & 73.58 & 73.75 & 62.99 \\ 
     & KDGQA & 91.61 & 71.35 & 71.34 & 59.60 \\ 
     & DGQA & 91.91 & 73.42 & 73.29 & 60.98 \\ 
     & PGQA & 90.93 & 72.06 & 72.02 & 60.65 \\

     \midrule
     
    \multirow{5}{*}{ViT-L} 
     & MHA & 96.08 & 84.97 & 83.24 & 78.19 \\ 
     & GQA & 94.05 & 76.61 & 76.46 & 65.62 \\ 
     & KDGQA & 87.33 & 61.24 & 66.85 & 54.33 \\ 
     & DGQA & \textbf{95.11} & \textbf{80.85} & \textbf{79.08} & \textbf{71.91} \\ 
     & PGQA & 92.71 & 70.37 & 72.61 & 65.20 \\
    
     \bottomrule
\end{tabular}
\caption{Test accuracies of the finetuning runs, post-uptraining on ImageNet-1k with a learning rate of $1\times 10^{-4}$, using variant-specific checkpoints. The models outperforming GQA in its category are highlighted in \textbf{bold}.}
\label{tab:results_1e-4_var_specific}
\end{table*}

\section{Impact of Uptraining Methodology}
\label{appendix_uptraining}

In our preliminary experiments, we explored the feasibility of reducing the computational cost of uptraining by utilizing the CINIC-10 dataset for model calibration, as opposed to the more resource-intensive ImageNet-1k. We uptrained on a variant of CINIC-10, which contained 270000 images, for $10$ epochs, which was just shy of double the amount of training steps required for $1$ epoch on ImageNet-1k.

The low-resolution, lack of variety in classes, and small size of the dataset was shown to be far from optimal for our models. For brevity, we elected to evaluate all models using the corresponding GQA checkpoints only, as can be seen in Table~\ref{tab:appendix_cinic10_results}.

\begin{table*}[h!]
\centering
\begin{tabular}{cccccc}
    \toprule
    
    \multirow{3}{*}{\textbf{Model}} & \multirow{3}{*}{\textbf{Variant}} & \multicolumn{4}{c}{\textbf{Accuracy}} \\
     &  & \textbf{CIFAR-10} & \textbf{CIFAR-100} & \textbf{Food101} & \textbf{Tiny ImageNet} \\
     
     \midrule

    \multirow{4}{*}{ViT-S} 
     & GQA & 98.13 & 72.78 & 61.83 & 53.88 \\ 
     & KDGQA & 95.01 & 62.71 & 53.83 & 44.61 \\ 
     & DGQA & 98.15 & 72.91 & 60.88 & 53.49 \\ 
     & PGQA & 97.68 & 66.42 & 54.02 & 48.37 \\

     \midrule

    \multirow{4}{*}{ViT-B} 
     & GQA & 96.74 & 67.36 & 56.46 & 49.26 \\ 
     & KDGQA & 95.23 & 61.28 & 50.14 & 41.74 \\ 
     & DGQA & 96.57 & 66.44 & 54.83 & 46.72 \\ 
     & PGQA & 96.25 & 63.52 & 50.34 & 44.41 \\

     \midrule
     
    \multirow{4}{*}{ViT-L} 
     & GQA & 97.95 & 77.31 & 66.80 & 59.85 \\ 
     & KDGQA & 97.08 & 71.38 & 60.17 & 54.37 \\ 
     & DGQA & 97.78 & 76.77 & 65.53 & 59.31 \\ 
     & PGQA & 97.77 & 74.11 & 63.08 & 56.77 \\
    
     \bottomrule
\end{tabular}
\caption{Test accuracies of the finetuning runs, post-uptraining on CINIC-10 with a learning rate of $1\times 10^{-5}$, using the GQA-only checkpoint.}
\label{tab:appendix_cinic10_results}
\end{table*}

Since CINIC-10 contains the same 10 classes from CIFAR-10, we find that the models obviously perform better on CIFAR-10. However, the shortcomings of the dataset are reflected when we move on to the remaining three datasets, where the models fail to generalize horrifically, experiencing significant drops in accuracy when compared to Table~\ref{tab:results_1e-5_gqackpt}.

This experiment shows that the quality of uptraining is very important, and that it was a necessary step to switch to ImageNet-1k. We believe that uptraining on ImageNet-21k and JFT-300M could improve the quality of the final fine-tuned models even more, but due to resource and time constraints, we were unable to explore this avenue.

\section{Head Similarity Bias}
\label{appendix_head_sim}
In MHA, the output of each head shows no dominant similarity with outputs from any other head. In fact, we see dissimilarity between outputs of a head compared to heads that are further away from it. 

However in GQA, output of heads are highly similar with other heads in the group causing an intra-group similarity bias which, among other reasons, might be contributing to a drop in accuracy in GQA compared to MHA. 

\begin{figure*}[h!]
    \centering
    \includegraphics[width=\columnwidth]{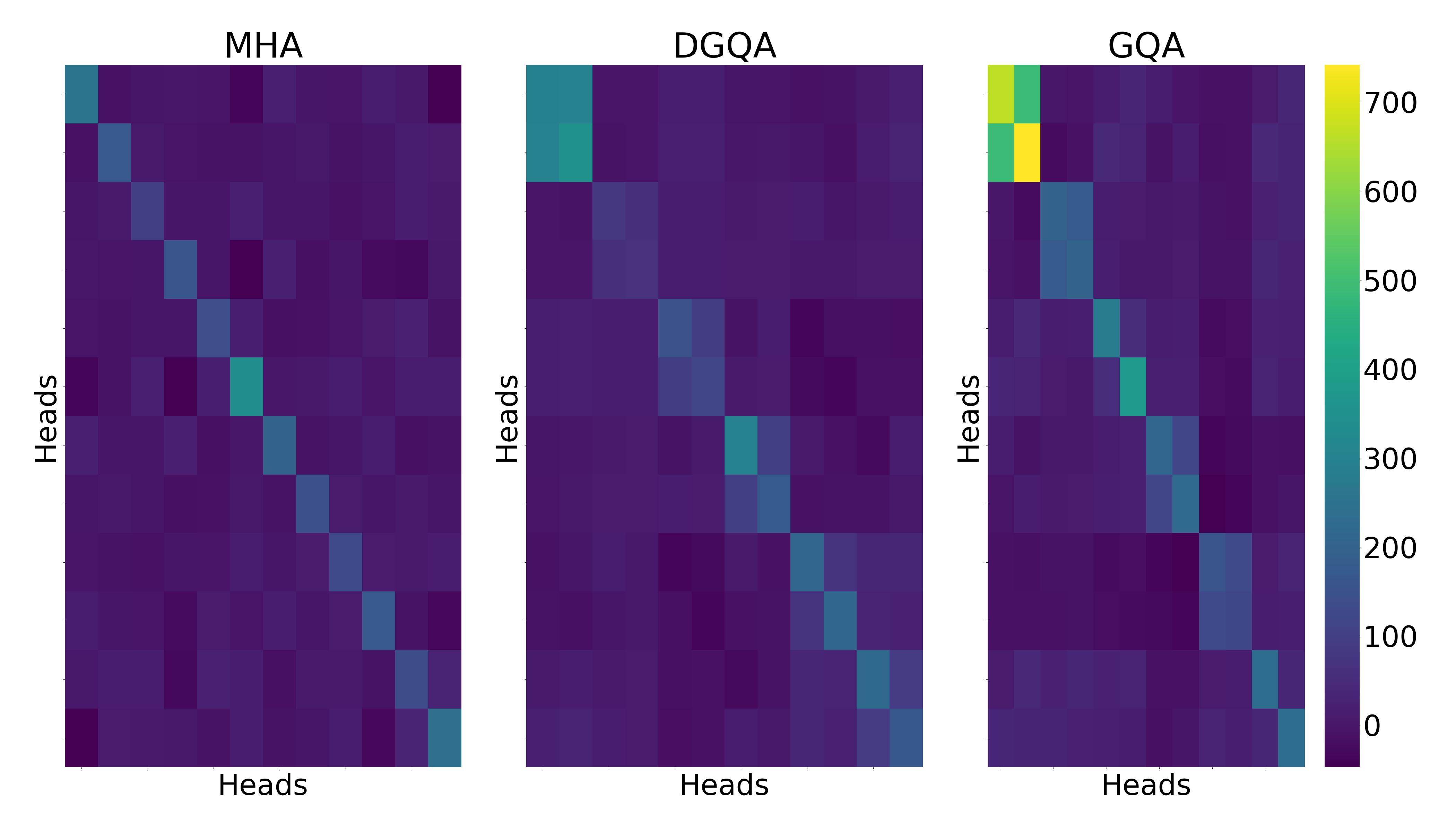}
    \caption{DGQA is an interesting middle ground between GQA and MHA. Unlike PGQA, heat map of DGQA does not wipe out patterns observed in MHA.}
    \label{fig:DGQA-heatmap}
\end{figure*}

\begin{figure*}[h!]
    \centering
    \includegraphics[width=\columnwidth]{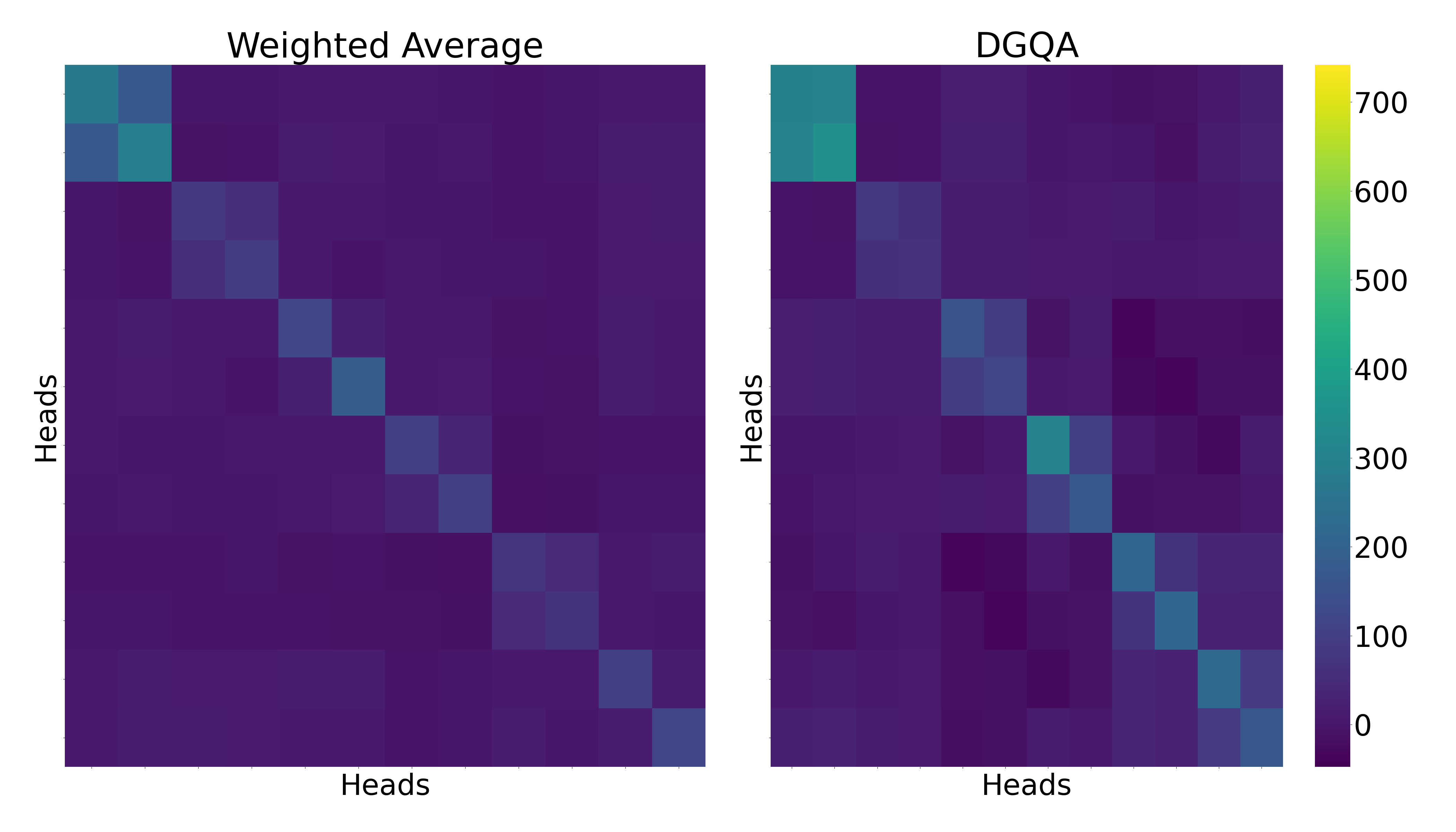}
    \caption{DGQA roughly approximates a weighted average between GQA and MHA. DGQA preserves trends from MHA while substantially alleviating the intra-group similarity bias afflicting GQA.}
    \label{fig:DGQA-avg}
\end{figure*}

We initially introduced PGQA as a way to mitigate this intra-group similarity bias, however we observed in Figure~\ref{fig:heatmap} that PGQA also wipes out other trends observed in the heat map of MHA. DGQA on the other hand proves to be an interesting middle ground between GQA and MHA (Figure~\ref{fig:DGQA-heatmap}). We show in Figure~\ref{fig:DGQA-avg} that the heat map of DGQA is close to a weighted average of GQA and MHA. This way, DGQA is able to largely mitigate the intra-group similarity bias while also retaining dissimilarity trends between heads as seen in heat map of MHA.

\section{Complex Data Needs more Non-Uniform Groups}
\label{appendix_groups_dist}

\begin{figure}[ht]
    \centering
    \includegraphics[width=\columnwidth]{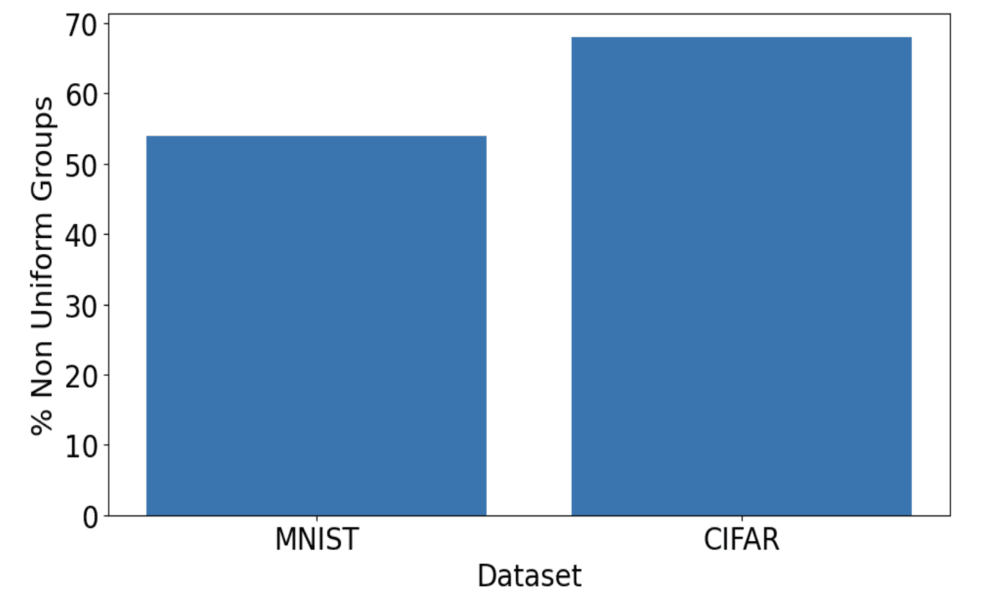}
    \caption{Model architecture being the same, it seems like the more complicated CIFAR10 prefers more non-uniform groups during training.}
    \label{fig:counts-dist}
\end{figure}

\begin{figure}[ht]
    \centering
    \includegraphics[width=\columnwidth]{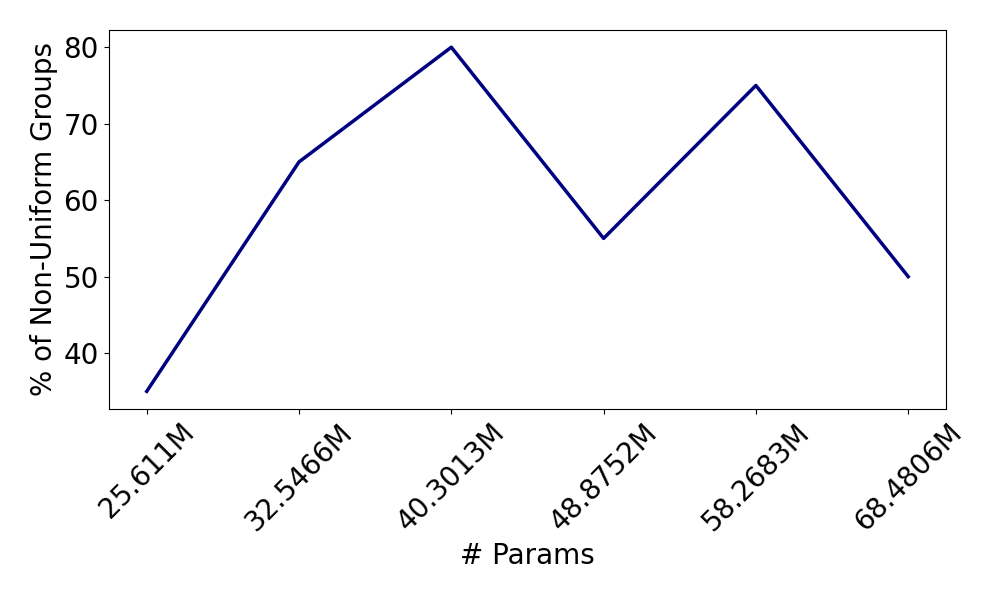}
    \caption{The $\%$ of non-uniform groups show no clearly pattern when we increase parameters while keeping the layers constant.}
    \label{fig:params-dist}
\end{figure}

\begin{figure}[h]
    \centering
    \includegraphics[width=\columnwidth]{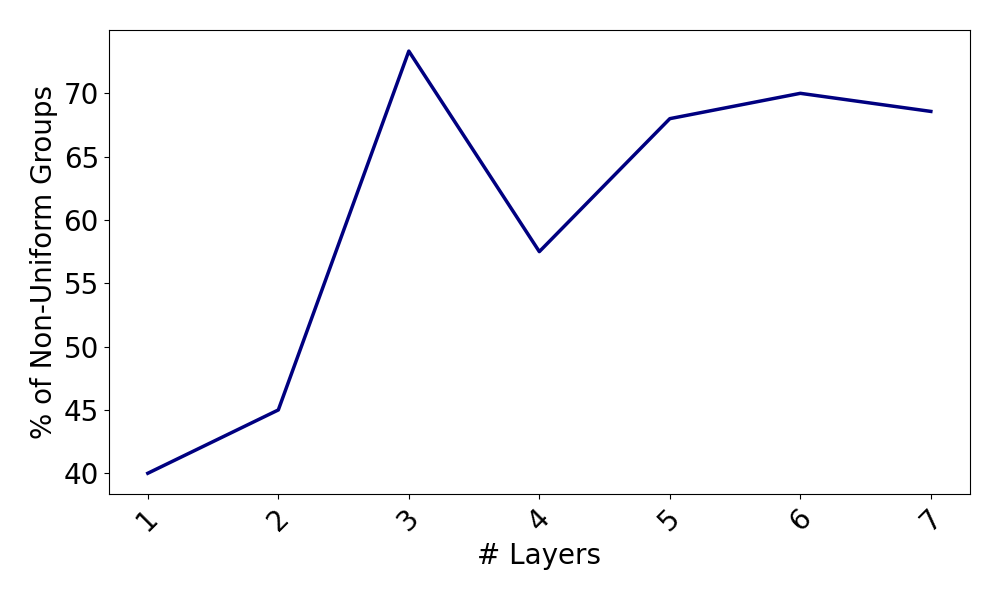}
    \caption{The $\%$ of non-uniform groups increase as we increase the number of layers in the model.}
    \label{fig:layers-dist}
\end{figure}

We trained the same ViT from scratch with the same hyper parameters (Batch Size of 32, a learning rate of $1e-4$, AdamW optimiser, and window size of 100) for 1000 training steps on MNIST and CIFAR-10 datasets. Our experiments in Figure \ref{fig:counts-dist} show that the more complex, feature dense CIFAR-10 dataset chooses a higher proportion of its allocations (which change every 100 training steps) to be non-uniform.

This simple experiment shows the dependence of group allocations on the dataset, however the static nature of GQA does not allow for these dynamic allocations which are provided by our DGQA mechanism.

\section{Correlation between Number of Parameters and Non-Uniform Groups}
\label{appendix_params_dist}

We repeated the experiments from Section \ref{appendix_groups_dist} in the appendix but we progressively increase number of parameters by increasing the embedding dimensionality but keeping the number of layers constant. Figure \ref{fig:params-dist} shows that while number of layers remain constant, simply increasing parameters does not necessarily lead to an increase in the number of non-uniform groups relative to total group allocations.

However, when we repeat these experiments again but this time, we keep the embedding dimensionality constant but increase the number of layers and hence the number of parameters too. Figure\ref{fig:layers-dist} shows a much stronger upward trend depicting that increasing the number of layers leads to a larger percentage of non-uniform group allocations.

\end{document}